\documentclass{article} 
\usepackage{iclr2021_conference,times}

\usepackage{amsmath,amsfonts,bm}

\def\eqref#1{equation~\ref{#1}}

\def\1{\bm{1}}

\DeclareMathAlphabet{\mathsfit}{\encodingdefault}{\sfdefault}{m}{sl}
\SetMathAlphabet{\mathsfit}{bold}{\encodingdefault}{\sfdefault}{bx}{n}

\usepackage{hyperref}
\usepackage{url}

\usepackage{times}
\usepackage{epsfig}
\usepackage{graphicx, xcolor}
\usepackage{amsmath}
\usepackage{amssymb}

\usepackage{graphicx, xcolor}
\usepackage{amsmath}
\usepackage{amssymb}
\usepackage{algorithmic}
\usepackage{caption}

\usepackage{times}
\usepackage{helvet}
\usepackage{courier}
\usepackage{lipsum}

\usepackage{bm}
\usepackage{color}
\usepackage{subfig}
\usepackage[linesnumbered,ruled,vlined]{algorithm2e}

\usepackage{array}
\usepackage{multirow}
\usepackage{booktabs}
\usepackage{rotating}

\usepackage{dcolumn}

\usepackage{amsthm}
\usepackage[utf8]{inputenc}
\usepackage[english]{babel}

\title{Adversarial training may be \\ a double-edged sword}

\author{Ali Rahmati$^*$, Seyed-Mohsen Moosavi-Dezfooli$^\dagger$, and Huaiyu Dai$^*$  \\
$^*$North Carolina State University,
Raleigh, NC \\ 
$^\dagger$Institute for Machine Learning,
ETH Z\"urich\\

\texttt{\{arahmat, hdai\}@ncsu.edu}\\
\texttt{seyed.moosavi@inf.ethz.ch}
}

\iclrfinalcopy 
\begin{document}

\maketitle

\begin{abstract}
Adversarial training has been shown as an effective approach to improve the robustness of image classifiers against white-box attacks.
However,
its effectiveness against black-box attacks is more nuanced. 
 In this work, we demonstrate that some geometric consequences of adversarial training on the decision boundary of deep networks give an edge to certain types of black-box attacks. In particular,
we define a metric called \textit{robustness gain} to  show that while adversarial training is an effective method to dramatically improve the robustness in  white-box scenarios, it may not provide such a good robustness gain against the more realistic decision-based black-box attacks. Moreover, we show that even the minimal perturbation white-box  attacks   can converge faster against adversarially-trained neural networks  compared to the regular ones.
\end{abstract}

\vspace{-2mm}
\section{Introduction}

Adversarial examples  can be crafted with intentionally designed  perturbations added to the inputs to fool deep neural network classifiers.
Adversarial attacks may be executed  in different categories depending on the attacker's level of information, including the white-box setting~\citep{carlini2017towards, goodfellow2014explaining, moosavi2016deepfool}, the score-based black-box setting~\citep{chen2017zoo, ilyas2018black,narodytska2016simple} and the decision-based black-box setting~\citep{ hsja, signopt, qfool,geoda}. 
 Generally, in the white-box scenario, the effectiveness of the attack is measured by the norm of minimal perturbations required to fool the network, while in the black-box settings, norm of  perturbations for a given number of queries is evaluated. Moreover, among the adversarial attacks in the literature, the most challenging ones are decision-based attacks, in which the attacker only has access to the output label of the image classifier for a given input.
On the other hand, adversarial training has been shown to be an effective method to improve the robustness of the image classifiers against adversarial attacks in the white-box setting~\citep{madry2018towards, shafahi2019adversarial,wong2019fast}. However, the effectiveness of adversarial training is not fully understood in the more practical real-world black-box settings. Therefore, the goal of this paper is to study the effectiveness of adversarial training in improving the robustness of image classifiers against decision-based black-box attacks where the attacker's level of information from the deep neural network image classifier is the least. 




In order to craft an adversarial perturbation in a decision-based black-box setting, the critical information is mostly the normal vector to the decision boundary of the image classifier. Obviously, the less information available to the attacker (which is more realistic), the less successful attack can be done. 
The estimation of the normal vector to the boundary in this setting  is conducted  with carefully designed  queries on the boundary of  the image classifier with the underlying assumption that the decision boundary has low mean curvature in the vicinity of inputs~\citep{hsja,signopt,qfool,geoda}. The goal of such black-box attacks  is to reduce the number of queries as much as possible with an efficient estimate of the normal vector to the boundary. Therefore, such estimators are expected to work better if the decision boundary is less curved as most of them rely on some sort of linearization of the decision boundary. Thus, a flatter boundary would be favorable to these types of attacks.

Interestingly, it is empirically shown that adversarial training leads to neural networks with flatter decision boundaries, compared to the boundaries learned by regular training methods~\citep{NEURIPS2019_0defd533, cure}. We will show that such feature of the adversarially-trained networks is indeed favorable for black-box attacks. The goal of this paper is to show some evidence that although the adversarial training improves the robustness of deep image classifiers effectively (i.e., it increases the minimum distance of datapoints to the boundary)
against the minimal norm perturbation white-box attacks, it becomes less effective in more practical attack settings. That is, decision-based black-box attacks can exploit the excessive flatness caused by adversarial training.
In particular, we define a metric called \textit{robustness gain} as the ratio of $\ell_2$ norm of adversarial perturbation  required to fool  the adversarially-trained network to that for the regular network. 
We observe that the robustness gain  against the  real-world attacks in which the least amount of information is available to the attacker is not as good as the one in the ideal white-box scenario with full information. 
 We also show that even iterative white-box attacks may converge faster against adversarially-trained networks due to their flatter boundary and more linear behaviour of  such networks.

 The rest of the paper is organized as follows. In Section~\ref{black}, we study the effectiveness of adversarial training  against decision-based black-box attacks.
In Section~\ref{white}, we evaluate the white-box attack  performance against adversarially-trained networks.  Finally, in Section~\ref{conc}, we conclude the paper and provide the direction forward.

 \vspace{-3mm}
\section{Adversarially-trained networks and decision-based black-box attacks}\label{black}
Although it is known that adversarial training is a quite effective approach to make image classifiers robust in the white-box settings, their performance in the more practical settings is not well investigated.
In this section, our goal is to evaluate the effectiveness of adversarial training against decision-based black-box attacks in which the attacker has only access to the output label of the image classifier for a given input. We  consider the most challenging setting in which only the top-1 label of the deep classifier is available to the attacker. Generally, such attacks aim to obtain the $\ell_2$-minimized norm perturbation with the minimum number of queries.

The most successful decision-based black-box attacks in the literature~\citep{ hsja, signopt, qfool,geoda} aim to estimate the normal vector to the decision boundary with a local linearization of the decision boundary at a randomly obtained boundary point. Such estimators rely on the fact that the boundary of the image classifiers has a low mean curvature in the vicinity of the data samples. As a result, the less curved the boundary is, the more valid such an assumption is.
An important observation regarding adversarial training~\citep{madry2018towards} is that, in addition to imposing a larger distance between the data point
and the decision boundary (hence resulting in a higher robustness),  the decision regions of adversarially trained
networks get flatter and more regular~\citep{NEURIPS2019_0defd533}. In particular,
 the curvature of the decision boundary decreases after adversarial training \citep{cure}. 
 It is  intuitive that a normal vector estimator will provide  a better estimation on smoother boundaries regardless of the boundary point at which the estimation is done. In particular, the flatter the boundary the more aligned the directions of the normal vectors at different boundary points are. In the extreme case that the boundary is a hyperplane, the normal vectors to the boundary  are in the same direction throughout  the hyperplane. Thus, interestingly, this results in a better estimation of the normal vector to the boundary for a adversarially-trained classifier in the black-box setting. This interesting observation might be exploited as a future direction   to develop new attacks with potentially a smaller number of queries to estimate the normal vector leading to a more effective attack. 
 
Although   adversarial training provides a quite impressive robustness against attackers by increasing the distance of the input to the boundary in the white-box setting, our goal is to see how such neural networks behave in query-limited black-box settings.  We define a metric called  \textit{robustness gain} $\eta =  \ell_2^\textrm{robust}/\ell_2^\textrm{regular}$ as the ratio of $\ell_2$ norm of the adversarial perturbation required to fool the adversarially-trained network to that for the regular network.
 We  conduct the experiments on a pre-trained ResNet-50~\citep{he2016deep} called the \textit{regular} network and the adversarially-trained ResNet-50~\citep{madry2018towards} called the \textit{robust} network throughout the paper. We consider a  random set   of 300 correctly classified images by both networks from the ILSVRC2012's validation set~\citep{deng2009imagenet}.

\vspace{-4mm}
\paragraph{Black-box attacks performance evaluation}

\begin{figure}[t!]
 \vspace{-5mm}
  \subfloat[\label{iterreg}]{%
      \includegraphics[ width=6 cm,height=3.5cm]{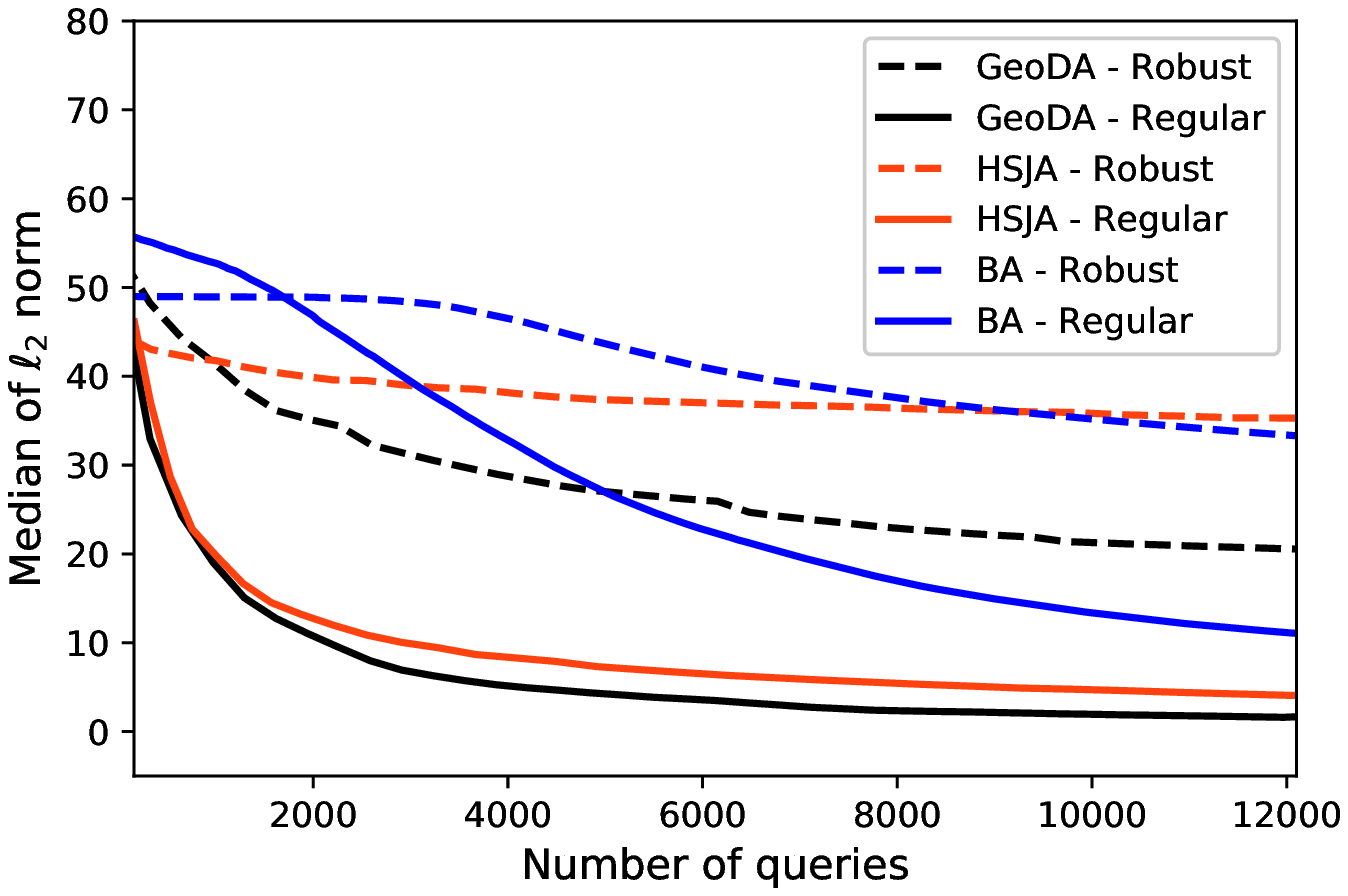}}
      \hspace{\fill}
  \subfloat[\label{iterrbst}   ]{%
      \includegraphics[ width=6 cm,height=3.5cm]{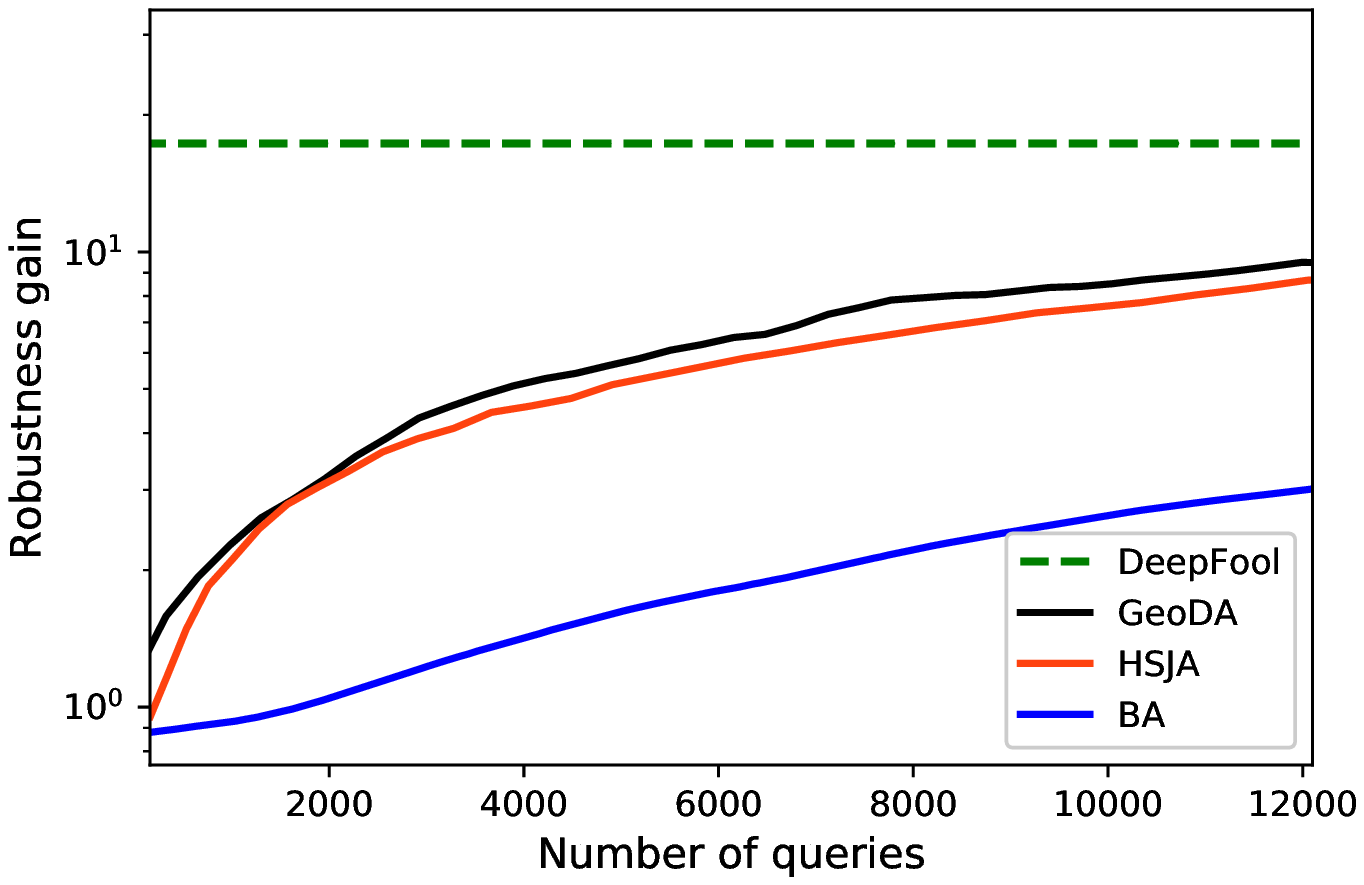}}\\
        \vspace{-3mm}
\caption{(a) Performance comparison of  different black-box  attacks for both regular and adversarially-trained ResNet50. (b) The robustness gain for $\ell_2$ norm under different attack scenarios. }\label{tab1}
\end{figure}

We compare the performance of black-box attacks  HSJA  \citep{hsja}, GeoDA~\citep{geoda},  boundary attack (BA)~\citep{brendel2018decisionss} on both regular and adversarially-trained ResNet-50 networks in Fig.~\ref{iterreg}. As shown, for a given query budget, the $\ell_2$ norm of perturbations for the attacks against the adversarially-trained network is larger compared to that of the regular network as expected. However, an interesting observation is that  while GeoDA  has almost the same $\ell_2$ norm  as HSJA for the regular network, it provides much smaller $\ell_2$ norm for perturbations against the adversarially-trained network compared to HSJA for a fixed amount of query budget. The main reason for this phenomenon is that GeoDA is \textit{explicitly} built based on the assumption that the boundary of the classifier has a low mean curvature. On the other hand, adversarially trained-networks has flatter decision boundaries which actually gives an edge to GeoDA. Thus, to attack   adversarially-trained networks more efficiently, it is beneficial for the attackers to deploy attacks    exploiting the flatness of the decision boundary.

\vspace{-2mm}
\paragraph{Robustness gain} In this part, our goal is to evaluate how much adversarially-trained networks can improve the robustness under various kind of attacks. We plot the \textit{robustness gain} for different attacks in Fig.~\ref{iterrbst}.  The larger the $\eta$ for a given attack is, the better the adversarial training  can improve the robustness  compared to the case of the regular network. 
In Fig.~\ref{iterrbst}, it is observed that $\eta$ is equal to approximately 17 (see Table~\ref{dfdf}) for the white-box attack DeepFool (DF)~\citep{moosavi2016deepfool} which is a quite good improvement. However, in the case of decision-based black-box setting,  first,  it can be seen that the robustness gain  is lower than that of the white-box attack in general. Second, by extracting more information from the image classifier (getting more queries), the robustness gain increases.  It  implies that the less information attacker knows about the image classifier, i.e., the more practical the attack scenario is (e.g., from white-box to black-box, or by decreasing the number of queries in the black-box setting), the less the adversarially-trained network can improve the robustness. That being said, adversarial training for the deep image classifiers is much more  effective against white-box scenarios than against black-box scenarios.

\vspace{-2mm}

\section{ Adversarially-trained networks and minimal norm ~~
white-box attacks}\label{white}
We already discussed  the effectiveness of the adversarially-trained network with respect to  the number of required queries against decision-based black-box attacks in the query-limited regime. In the white-box scenario, we evaluate the effectiveness through  the number of required iterations for the convergence of a minimal $\ell_2$ norm  perturbation white-box attack. To this end, we choose a minimal $\ell_2$ norm white-box attack  DeepFool~\citep{moosavi2016deepfool} and compare its performance on an adversarially-trained \citep{madry2018towards} and a regular ResNet-50 in Table~\ref{dfdf}. 
    The main reason we choose DeepFool is its dependence  on linearizing the output function of the classifier. The algorithm starts with  locally linearizing the output function of the classifier and repeats such an approximation  iteratively  to compensate for the effect of the non-linearity of the output function.  The more linear the output function of the image classifier is, the fewer iterations required for DeepFool to converge.  Interestingly, despite that the adversarially-trained network requires perturbations with larger $\ell_2$ norm to be fooled, due to the more linear behavior of its output function, DeepFool converges faster on this network. In this sense, one can conclude that there is a trade-off to attack adversarially-trained networks even in the white-box setting.

\begin{table} [h]
\centering 
\begin{tabular}{c | c |c| c c c  } 
\toprule 

 & \small{\textbf{Median of Iterations}} & \small{\textbf{Max of Iterations}} &\small{\textbf{  $\ell_2$ norm  }}     \\

\midrule  \midrule
Regular  \citep{he2016deep}& 4 & 15 &   0.209  \\
  \midrule 
Adv. trained \citep{madry2018towards} & 2&  4 & 3.618  \\
\bottomrule
\end{tabular}
\caption{ DeepFool \citep{moosavi2016deepfool} performance on adversarially-trained and regular ResNet-50 networks (over 1000 correctly classified  samples from ImageNet validation set).} 
\label{dfdf} 
\end{table}

\begin{figure}[t!]
 \vspace{-5mm}
  \subfloat[\label{f21}]{%
      \includegraphics[ width=6 cm,height=3.5cm]{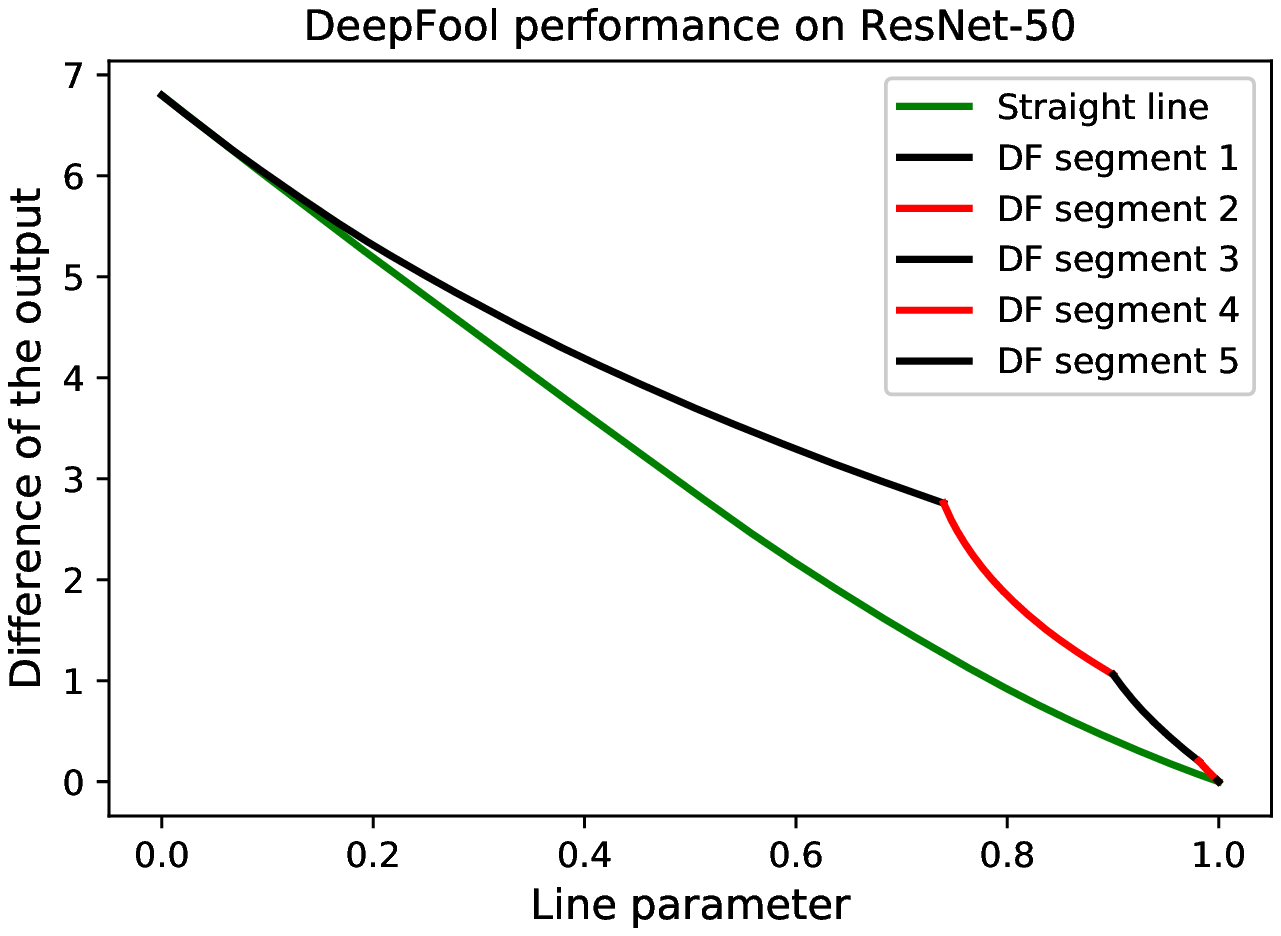}}
      \hspace{\fill}
  \subfloat[\label{f22}   ]{%
      \includegraphics[ width=6 cm,height=3.5cm]{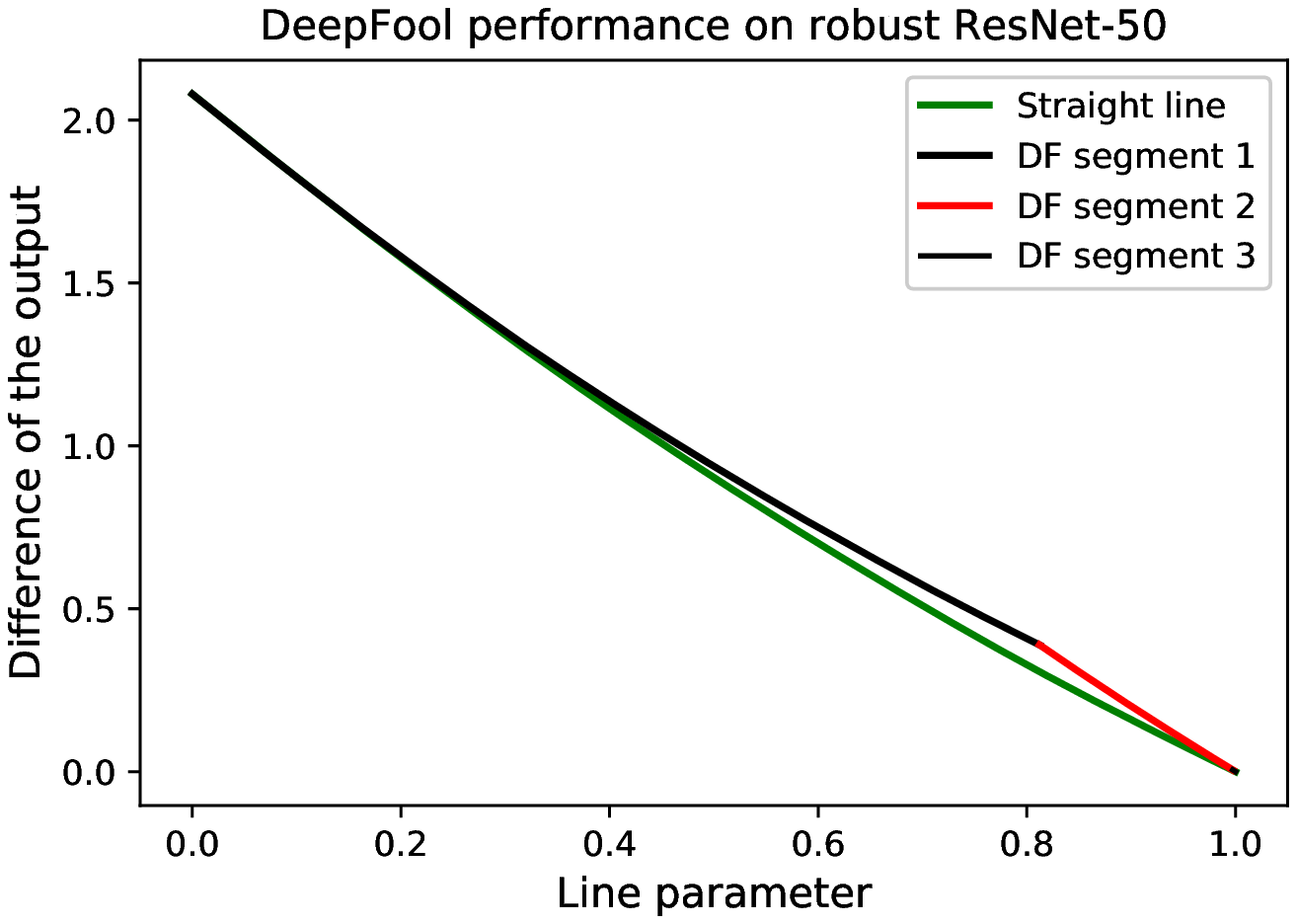}}\\
        \vspace{-3mm}
\caption{Performance evaluation of DeepFool over different iterations on (a) Regular ResNet 50 network. (b) Adversarially-trained ResNet 50 network. }\label{tab1}
\end{figure}

In the next experiment, the goal is to qualitatively study the behaviour of output function along the trajectory of the iterations of DeepFool for a single data point. 
In this case, DeepFool requires 3 and 5 iterations to converge for adversarially-trained and regular  ResNet-50, respectively. {We consider the difference of the logits corresponding to the original and the adversarial labels for our evaluation. We  track this difference along two paths: 1) the straight path between the original image and the DeepFool  adversarial  example (i.e., green line in Figs.~\ref{f21} and ~\ref{f22}), and 2) the path taken by DF in each iteration (i.e. black and red line segments). We generate  images on the line from the original image to the minimal perturbation adversarial example obtained by DeepFool.
 By varying the line parameter $t$, we consider the images along the line $\bm{x}= \bm{x}_0 + t(\bm{x}_{adv}-\bm{x}_0)$, where $t=0$ corresponds to the original image and $t=1$ gives the adversarial image which falls on the boundary. When the image is on the clean label side, the output value of the clean label is larger than the adversarial label. Approaching the boundary, this difference decreases where on the boundary the difference is equal to zero and the transition occurs. Assuming $\bm{x}_i$ as the output of DeepFool  in iteration $i$, each line segment $i$ (i.e. black and red segments in Figs.~\ref{f21} and ~\ref{f22}) is corresponding to the images on the line $\bm{x}= \bm{x}_{(i-1)} + t(\bm{x}_i-\bm{x}_{(i-1)})$ for $t\in[0,1]$, where  $\bm{x}_i = \bm{x}_{adv}$ if $i$ is the last iteration.} First, it can be seen that the straight path (green line) is much closer to the path constructed with DeepFool iterations for the adversarially-trained network compared to that of regular network.  Second, it is shown that even in each line segment corresponding to each iteration traversed by DeepFool algorithm, there is more non-linearity in regular networks. As a result, it can be seen that although  adversarial training improve the robustness (increases the minimal $\ell_2$ norm required for successful attack), it gives an edge to  the attacker  (with a smaller number of iterations to converge) due to more linear behaviour of adversarially-trained networks.

\vspace{-3mm}

\section{Conclusion and Direction forward}\label{conc}
\vspace{-2mm}
We showed that although  adversarial training is quiet effective against white-box  attacks,  in query-limited decision-based black-box attacks,  it may not perform as efficiently as in the case for the white-box attacks.
We demonstrated that since  adversarial training leads to a significantly flatter boundary and a more linear behavior of the image classifier, it can give an edge to certain types of black-box attackers whose goal is to estimate the normal vector to the boundary. 
This feature of the adversarially-trained networks can also provide a chance for minimal norm perturbation whit-box attacks to produce adversarial examples with a smaller number of iterations. As a future direction, such properties of adversarially-trained networks can be exploited, especially in the black-box settings, to design more effective attacks against such networks.

\bibliography{egbib}
\bibliographystyle{iclr2021_conference}


\end{document}